\documentclass[conference]{IEEEtran}
\usepackage{cite}
\usepackage{amsmath,amssymb,amsfonts}
\usepackage{algorithmic}
\usepackage{graphicx}
\usepackage{textcomp}
\usepackage{xcolor}
\def\BibTeX{{\rm B\kern-.05em{\sc i\kern-.025em b}\kern-.08em
    T\kern-.1667em\lower.7ex\hbox{E}\kern-.125emX}}

\usepackage{algorithm}

\usepackage{amssymb}

\usepackage{comment}

\usepackage{amsthm}

\usepackage{subfig}
\usepackage{caption}

\usepackage{mathtools}

\usepackage{bm}

\usepackage{array}
\usepackage{booktabs}
\usepackage{multirow}

\newcommand{\midsepremove}{\aboverulesep = 0mm \belowrulesep = 0mm}
\newcommand{\midsepdefault}{\aboverulesep = 0.605mm \belowrulesep = 0.984mm}

\usepackage{colortbl}
\definecolor{Green}{rgb}{0.7,1,0.7}
\definecolor{Yellow}{rgb}{1,1,0.7}
\definecolor{Red}{rgb}{1,0.8,0.8}

\newtheorem*{defi}{Definition}

\begin{document}

\title{An Optimization Framework for Task Sequencing in Curriculum Learning}

\author{\IEEEauthorblockN{Francesco Foglino}
\IEEEauthorblockA{\textit{School of Computing}\\
\textit{University of Leeds}\\
Leeds, United Kingdom \\
scff@leeds.ac.uk}
\and
\IEEEauthorblockN{Christiano Coletto Christakou}
	\IEEEauthorblockA{\textit{School of Computing}\\
		\textit{University of Leeds}\\
		Leeds, United Kingdom \\
		mm18ccc@leeds.ac.uk}
\and
\IEEEauthorblockN{Matteo Leonetti}
\IEEEauthorblockA{\textit{School of Computing}\\
\textit{University of Leeds}\\
Leeds, United Kingdom \\
m.leonetti@leeds.ac.uk}
}

\maketitle

\begin{abstract}
Curriculum learning in reinforcement learning is used to shape exploration by presenting the agent with increasingly complex tasks. The idea of curriculum learning has been largely applied in both animal training and pedagogy. In reinforcement learning, all previous task sequencing methods have shaped exploration with the objective of reducing the time to reach a given performance level. We propose novel uses of curriculum learning, which arise from choosing different objective functions. Furthermore, we define a general optimization framework for task sequencing and evaluate the performance of popular metaheuristic search methods on several tasks. We show that curriculum learning can be successfully used to: improve the initial performance, take fewer suboptimal actions during exploration, and discover better policies.
\end{abstract}

\begin{IEEEkeywords}
Curriculum Learning, Transfer Learning, Reinforcement Learning
\end{IEEEkeywords}

\section{Introduction}
Learning through a sequence of increasingly difficult tasks is a technique employed by humans in order to approach extremely complex domains. Examples range from learning how to play musical instruments to sports, and encompass human education from primary school to university. This concept was proven to be extremely effective in biology as well, where experiments resulted in animals manifesting behaviors that members of their species had never shown before. It is not the case that their ancestors were incapable of such behaviors, but, as Skinner put it: ``nature had simply never arranged effective sequences of schedules'' ~\cite{skinner1958reinforcement, peterson2004day}.

Curriculum learning has been used in Reinforcement Learning (RL) in both software agents~\cite{shao2018micromancl,doom2017} and robots~\cite{learningbyPlaying2018,panzner2017deep}, to let the agent
progress more quickly towards better behaviors. Initial attempts have been made in automating curriculum generation, both in terms of generating appropriate tasks for the curriculum~\cite{narvekar16}, and sequencing them optimally~\cite{florensa2018goalgen,narvekarIJCAI17,leonetti17}. Experience sequencing is a central theme of curriculum learning, as the learning paradigm can entirely been seen as presenting the agent with the right task at the right time in its development.

Task sequencing algorithms have been developed to create an optimal curriculum in terms of training time~\cite{leonetti17,daSilva2018,narvekarIJCAI17}, that is, curriculum learning has been used as a technique to learn \emph{faster}. By learning through the curriculum, and transferring knowledge from task to task, the agent achieves a given level of performance more quickly then by learning the final, and most difficult, task directly. 

\begin{figure}[h!]
	\includegraphics[width=\linewidth]{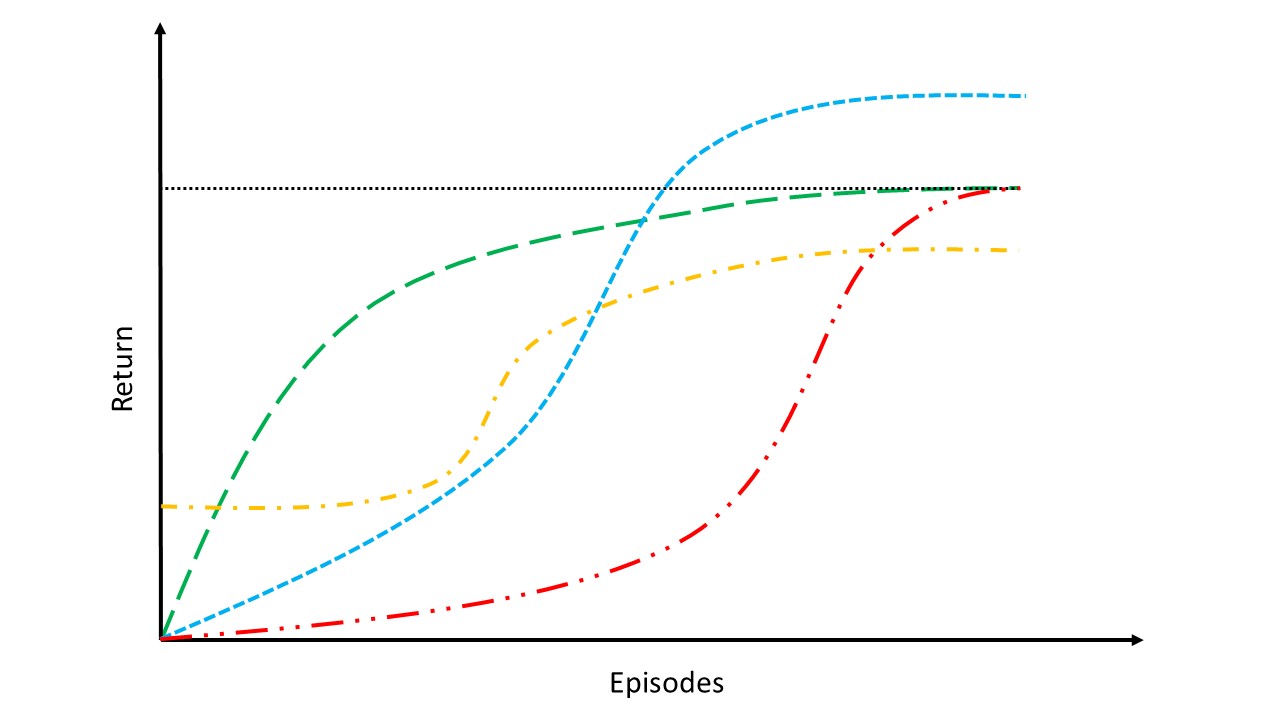}
	\caption{Illustration of different objectives for curriculum learning}
	\label{Fig:metricsillustration}
\end{figure}

Figure \ref{Fig:metricsillustration} is an illustration of different possible learning curves in a task of interest. The red and green line take the same time to reach a given threshold, but the green one accumulates more reward during learning. The yellow line starts with a better initial performance than the others, and the blue line finds a better  policy by the end. Existing algorithms have focused on shortening the time required by the red line to reach the threshold.

In this paper, we define a more general formulation of the sequencing problem, and introduce two novel uses of task sequencing for curriculum learning: exploring more efficiently in a critical task, and discovering better policies in extremely complex tasks. We  describe motivating scenarios for both cases, in which the total training time is not the main concern. As a consequence, we adopt different objective functions, which focus the benefit of the curriculum on novel aspects of the learning process. In our illustration, this corresponds to using a curriculum to increase the initial \emph{jumpstart}, like in the yellow line; accumulating more reward during learning, like the green line; or discover a higher-quality policy, like the blue line. We show that curriculum learning can be successfully leveraged with these novel objectives, obtaining significant improvements over learning from scratch. Furthermore, we identify the most suitable class of optimization algorithms for the introduced general problem formulation, and carry out an extensive experimental evaluation, in order to establish the most appropriate one for task sequencing with different objectives.

\section{Background}

\subsection{Reinforcement Learning}
We model tasks as episodic Markov Decision Processes. An MDP is a tuple $\langle S,A,p,r,\gamma \rangle$, where $S$ is the set of states, $A$ is the set of actions, $p : S \times A  \times S \rightarrow [0,1]$ is the transition function, $r: S \times A \rightarrow \mathbb{R}$ is the reward function and $\gamma \in [0, 1]$ is the discount factor. If a state is represented as a vector $s = \langle v_1, \ldots, v_d \rangle$ of $d$ variables the representation of the state space is said to be \emph{factored}. Episodic tasks have \emph{absorbing} states, that are states that can never be left, and from which the agent only receives a reward of $0$.

For each time step $t$, the agent receives an observation of the state and takes an action according to a policy $\pi : S \times A \rightarrow [0,1]$. The aim of the agent is to find the \textit{optimal} policy $\pi^*$ that maximizes the expected discounted return $G_0 = \sum_{t=0}^{t_M} \gamma^t r(S_t, A_t)$, where $t_M$ is the maximum length of the episode.

Sarsa($\lambda$) is a learning algorithm that takes advantage of an estimate of the \emph{value} function $q_\pi(s,a) = E_\pi [G_t \mid S_t = s, A_t = a]$. We represent the value function with a linear function approximator, so that the learning algorithm computes an estimate $\hat{q}(s,a) = \bm{\theta}^T \bm{\phi}(s,a)$ of $q_\pi(s,a)$ as a liner combination of \emph{features} $\bm{\phi}$.

\subsection{Transfer Learning}
Curriculum learning leverages transfer learning to transfer knowledge through the curriculum, in order to benefit a final task. Transfer takes place between pairs of tasks, referred to as the \emph{source} and the \emph{target} of the transfer. We use a transfer learning method based on value function transfer~\cite{tl4rl},  which uses the learned source q-values, representing the knowledge acquired in the source task,  to initialize the value function of the target task. Several metrics have been designed to evaluate transfer learning~\cite{tl4rl}, and we employ three of them for curriculum learning: time-to-threshold, jumpstart, and asymptotic performance.

\subsection{Combinatorial Optimization}
\label{sec:combinatorialop}
Combinatorial Optimization (CO) problems are characterized by the goal of finding the optimal configuration of a set of discrete variables. The most popular approaches in this field, called \emph{metaheuristics}, are \emph{approximate} algorithms, that do not attempt to search the solution space completely, but give up global optimality in favor of finding a good solution more quickly. Metaheuristics are applicable to a large class of optimization problems, and are the most appropriate methods for black-box combinatorial optimization, when a particular structure of the objective function (for instance, convexity) cannot be exploited. Task sequencing is one such black-box problem, therefore we selected four of the most popular metaheuristics algorithms for comparison with our search method: Beam Search \cite{lowerre1976harpy,ow1988filtered}, Tabu Search \cite{glover1989tabu}, Genetic Algorithm \cite{goldberg1989genetic}, and Ant Colony Search \cite{dorigo1991ant}. Beam Search and Tabu Search are \emph{trajectory based} algorithms, which starting from a single instance, search through the neighborhood of the current solution for an improvement. Genetic Algorithm and Ant Colony Search are \emph{population based} algorithms, that start from a set of candidate solutions, and improve them iteratively towards successive areas of interest.

\section{Related Work}

Curriculum Learning  in reinforcement learning is an increasingly popular
field, with successful examples of applications in first-person
shooter games \cite{doom2017,wu2018}, real-time strategy games \cite{shao2018micromancl}, and real-world robotics applications~\cite{learningbyPlaying2018}.

Curriculum Learning has initially been introduced in supervised learning \cite{bengio2009}. Closely related fields are multi-task reinforcement learning \cite{wilson2007multi}, and lifelong learning \cite{ruvolo2013ella}, where the agent attempts to maximize its performance over the entire set of tasks, which may not all be related to one another.  Conversely, in curriculum learning, the intermediate tasks are generated specifically to be part of the curriculum, and the curriculum is optimized for a single final task.

Task Selection has been studied for general transfer learning, and presents common aspects with the task selection that is part of sequencing in  curriculum learning. Several approaches consider learning a mapping from source tasks to target tasks, and estimating the benefit of transferring between the tasks \cite{ammar2014automated,sinapov2015learning,isele2016using}. Nonetheless, transfer learning is usually performed between two tasks, a source and a target, and task selection methods have never been leveraged to achieve longer sequences.

Curriculum learning has been used to guide the exploration by generating a sequence of goals or initial states within the same environment  \cite{asada1996purposive,sukhbaatar2018,florensa2018goalgen} where all tasks have the same dynamics. On the other hand, we consider the problem of sequencing entirely different tasks, in general between environments that differ in any of their MDP components.

The automatic generation of curricula has been divided into two sub-problems: task generation~\cite{narvekar16,daSilva2018}, that is the problem of creating a set of tasks such that transferring from them is most likely beneficial for the final task; and task sequencing~\cite{leonetti17,daSilva2018,narvekarIJCAI17}, whereby previously generated tasks are optimally selected and ordered. 

Current methods for task sequencing attempt to determine the optimal order of tasks
either with~\cite{narvekarIJCAI17,baranes2013active} or without
\cite{leonetti17,daSilva2018} executing the tasks. All current methods are heuristic approaches targeted at a particular objective: time-to-threshold~\cite{tl4rl}.

Task sequencing methods are the closest to our work, so we discuss them here in further detail.  Svetlik et al. ~\cite{leonetti17} propose a method to create a curriculum \emph{graph} without executing any task, therefore independently of the agent. The method is based on a manually specified task descriptor, and a heuristic measure of task similarity.  Da Silva and Costa ~\cite{daSilva2018} proposed an object-oriented extension of that method, and introduced a task generation and transfer learning procedure, also based on the object-oriented representation. Both methods do not explore the space of curricula, but directly build a single curriculum. As a consequence, they are the most efficient in terms of training time to build the curriculum, but also have no guarantee on the quality of the solution. Narvekar et al. ~\cite{narvekarIJCAI17} frame the curriculum learning problem as a higher-level MDP, where the state space is formed by all the possible policies, and the actions select the next task to learn. Such an MDP, however, is too complex to be solved directly, and the authors propose a greedy algorithm, guided by a measure of policy change. In its greedy selection, the algorithm prefers tasks that modify the policy the most. Differently from the works above, we consider several objective functions for task sequencing other than time-to-threshold, proposing a more general framework. Furthermore, we cast the problem as a black-box combinatorial optimization problem, which allows us to benefit from long-established metaheuristic search algorithms, with stronger quality guarantees.

\section{Task Sequencing Framework}
\subsection{Problem Definition}
\label{sec:probdef}
Let $\mathcal{M}$ be a set of MDPs, composed of $\mathcal{T\subset\mathcal{M}}$,
a finite set of candidate tasks for the curriculum, and $m_f \in \mathcal{M}$, a \emph{final} task. The final task is the task the designer wants the agent to learn more efficiently through the curriculum. We define a curriculum as a sequence of tasks in $\mathcal{T}$ without repetitions:
\begin{defi}{\emph{[Curriculum]}} Given a set of tasks $\mathcal{T}$,
	a \emph{curriculum} over $\mathcal{T}$ of length $l$ is a sequence
	of tasks $c=\langle m_{1},m_{2},$ $\ldots,$ $m_{l}\rangle$ where each $m_{i}\in\mathcal{T}$, and $\forall i,j \in [1,l] \; i \neq j \Rightarrow m_i \neq m_j$.
\end{defi}
Let $\mathcal{C}_{l}^{\mathcal{T}}$ be the
set of all curricula over $\mathcal{T}$ of length $l$. In the
rest of this paper we will drop the superscript wherever the set of candidate
tasks is implicit in the context.

We define $\mathcal{C}_{\leq L} \coloneqq \bigcup_{l=0}^L C_l$ as the set of all curricula of length at most $L$. We represent with $C_0$ the set containing the \emph{empty} curriculum of length $0$, denoted with $\langle \rangle$. The empty curriculum corresponds to learning the final task directly.
Given a performance metric $\mathcal{P}:\mathcal{C}_{\leq L}\times\mathcal{M}\rightarrow\mathbb{R}$,
which evaluates curricula for a specific final task, we consider the
problem of finding an optimal curriculum $c^{*}$, such that: 
\[
\mathcal{P}(c^{*},m_{f})\geq\mathcal{P}(c,m_{f})\hspace{0.5cm}\forall c\in\mathcal{C}_{\leq L}.
\]

\subsection{Performance Metrics}
\label{sec:metrics}

In this section we describe the objective functions we propose to use in our framework, and later assign them to new scenarios for novel uses of curriculum learning. 


Regret (Reg) is one of the metrics used to optimally balance exploration and exploitation in single-task learning, whereby the agent minimizing regret attempts to converge to the optimal policy while acting suboptimally as little as possible. The regret metric, with respect to a given performance threshold $g$ (which can be the value of the optimal policy when known), is defined as follows: 
\[
\mathcal{P}_{r}(c,m_{f}) \coloneqq -(Ng - \sum_{i=1}^{N}G_{f}^{i}),
\]
where $N$ is the number of episodes executed in the final task, and
$Ng-\sum_{i=1}^{N}G_{f}^{i}$ is the difference between the return
the agent would achieve if it obtained a return $g$ at each episode
and the return actually achieved. This difference is the \emph{regret}
with respect to a policy achieving $g$, and we intend to minimize it,
therefore it is multiplied by $-1$ since $\mathcal{P}_{r}$ is maximized.

The following objective functions have been defined in the context of single-task transfer learning~\cite{tl4rl, lazaric2012transfer}, and can be imported into curriculum learning.
Jumpstart (JS) evaluates the average reward of the agent within the first $D$ episodes:%
\[
\mathcal{P}_{j}(c,m_{f}) \coloneqq \frac{1}{D}\sum_{i=1}^{D}G_{f}^{i},
\]
where $G_{f}^{i}$ is the return obtained during episode $i$ in task
$m_{f}$.
Jumpstart can be used if it is crucial that the agent is deployed in the final task with the highest possible initial performance, and is a version of regret that focuses only a few initial episodes.

Jumpstart and regret are objectives over the quality of the exploration (how it starts, and how it proceeds respectively) in the final task. Hence, they are evaluated on every episode. In the following two metrics, we focus on the value of the learned policy rather than on exploration. The value of the actions taken during learning is not the objective of the optimization, and the agent aims at either reaching a certain performance level faster, or reach a higher level. For this reason, every $K$ learning episodes, we introduce an \textit{evaluation} phase, in which the current policy is executed $Q$ times with no exploratory actions, to estimate its expected return. During this phase, no updates are performed to the value function (or the policy). In the rest of this section, we denote with $\mathcal{E}$ the set of the evaluation steps.


The objective \emph{max-return} (MR) focuses on the value of the policy learned within a given horizon:
\[
\mathcal{P}_{m}(c,m_{f}) \coloneqq \max_{I \in \mathcal{E} }  G_{f}^{I},
\]
where $G_{f}^{I} = \frac{1}{Q} \sum_{i =1}^Q G_f^i$ is the average return over the $Q$ episodes in the evaluation step $I$ on the final task. This is conceptually equivalent to \emph{asymptotic performance} introduced for transfer learning~\cite{tl4rl}, whereby the agent maximizes its performance by the end of learning. Max-return takes into account the non-monotonic nature of learning with function approximation, so that the best discovered behavior may not be at the end of a trial, but anywhere during it.


Lastly, we consider what is currently the most used objective: time-to-threshold (TTT). It evaluates the number of actions executed throughout the curriculum in order to achieve a given threshold performance
$g$ during an evaluation step in the final task. Let $a(m_{i})$ be the number of actions the agent
executed in task $m_{i}$ before moving on to the next task in the
curriculum, and $a_{g}(m_f)$ be the number of actions the agent executed
in the final task until the evaluation step in which the policy achieves an average return of $g$.
The time-to-threshold metric is defined as follows: 
\[
\mathcal{P}_{t}(c,m_{f}) \coloneqq - (a_{g}(m_f) + \sum_{m_{i}\in c}a(m_{i})  ),
\]
where, similarly to the regret, we intend to minimize the time to
threshold, therefore the total time is multiplied by $-1$. Time-to-threshold is the only metric in which each task contributes to the total performance explicitly. In the other metrics the intermediate tasks affect the performance exclusively through transfer learning, and its effect on the behavior in the final task.

\subsection{Scenarios}

In MDPs, if the value function can be represented exactly, and the exploration strategy guarantees that every action is executed in every state enough many times, the value function converges to the value of the optimal policy, regardless of its initialization. In tasks of practical interest, however, both the use of functions approximators, and the need for a more limited exploration, determine convergence to a local optimum at best, which depends on the initial value. The role of the curriculum is to identify the optimal initial value, that is, to prepare the agent to the final task as best as possible.  In the rest of this section we describe three scenarios with their requirements, and in the rest of the paper we show how to apply the curriculum learning framework to them.

The first scenario is provided by \emph{critical} tasks, where exploration is costly and suboptimal actions must be limited as much as possible. Examples of this case are abundant in robotics, where limiting exploration is one of the main concerns. We assume the existence of a simulator, also a common occurrence for such domains, since learning cannot be performed in the real critical task directly. A final task $m_f$, modeling the real task, can be set up in simulation, and the optimal curriculum computed without the need to act in the real task. The objective functions used in this scenario are jumpstart and regret. Either one can be chosen separately, or in combination, to obtain that the agent starts with good performance and explores as efficiently as possible thereafter. At the time of deployment, the value function of the last task of the curriculum is used to initialize the agent in the real task, since, by construction, this is the best possible initialization to the model of the real task used for training.

The second scenario corresponds to \emph{complex} tasks. In tasks complex enough, the optimal policy is unknown, and the agent cannot be guaranteed to achieve it in any feasible amount of time. One example of such a case is the game StarCraft~\cite{shao2018micromancl}. In this scenario, exploration is not a concern, as long as the agent  achieves a policy of high value. Furthermore, training time is secondary, since the agent would take an enormous amount of time if not learning through the curriculum anyway. In this case, we are interested in the initial value function that can make the agent discover the best possible policy in the final task. The objective function for this case is max-return.

The third scenario is the one previously considered in the literature, and therefore we will only introduce it briefly. A large task is broken down into smaller subproblems, so that the agent can learn the optimal policy faster by learning the subproblems in sequence. The objective function for this scenario is time-to-threshold, the one commonly used in curriculum learning.

\section{Algorithms for Task Sequencing}

The objective functions defined in Section \ref{sec:metrics} do not have an explicit closed-form definition, since the actual return obtained by the agent can only be measured during learning. Therefore the resulting optimization problem is black-box, and it is in general nonsmooth, nonconvex, and even discontinuous.  Furthermore, the optimization problem is constrained to a combinatorial feasible set. The most appropriate class of optimization algorithms for this type of problem is the class of metaheuristc algorithms, introduced in Section \ref{sec:combinatorialop}. We selected four popular and representative algorithms from this class, two of which are trajectory-based, while the other two are population-based. In this section we describe the customization we performed to these otherwise general algorithms, to apply them for task sequencing. We set out to evaluate them experimentally, in order to determine which ones are the most appropriate for curriculum learning. 


As a simple baseline, we use a purely greedy algorithm. This is meant to be an easy and fast solution which could be the only available option for particularly complex environments. The greedy search starts from a set of candidate curricula composed by all the curricula of length $1$. These are then evaluated for selecting the best candidate. The best candidate is then given as input to \texttt{GenerateCandidates}, to obtain the next set of candidates. The algorithm terminates when the best candidate does not improve on the current best curriculum (hence its greedy nature). 
The function  \texttt{GenerateCandidates} is shown in Algorithm \ref{Alg:GenerateCandidates}.
\begin{algorithm}[h!]
	\caption{GenerateCandidates}
	\label{Alg:GenerateCandidates}
	\begin{algorithmic}[1]
		\REQUIRE \texttt{seeds}, $\mathcal{T}$
		\ENSURE candidate set $C$
		
		\STATE $C \leftarrow  \emptyset$ 
		\FOR{$c \in$ \texttt{seeds}}
		\STATE$E \leftarrow \{m_{i}\in\mathcal{T}\vert m_{i}\notin c\}$
		\FOR{ $m \in E$}
		\STATE append $m$ to $c$
		\STATE $C \leftarrow C \bigcup c$
		\ENDFOR
		\ENDFOR
		\RETURN $C$
	\end{algorithmic}
\end{algorithm}
The new set of candidates is computed by appending, to each of the \emph{seed} curricula, all tasks that do not already belong to that curriculum. For example, if $\mathcal{T} = \{m_1,  m_2 ,  m_3  \}$, and \texttt{GenerateCandidates} is invoked on $\{ \langle m_1 \rangle, \langle m_2 \rangle\}$, it returns $\{\langle m_1, m_2 \rangle, \langle m_1, m_3 \rangle, \langle m_2, m_1 \rangle \langle m_2, m_3 \rangle\}$.
Greedy search is the simplest deterministic local search algorithm, and is therefore easily prone to stop at a local maximum. Next we consider stochastic algorithms able to deal with more flexible definitions of locality.

In Beam Search \cite{lowerre1976harpy,ow1988filtered} we start from an empty sequence of source tasks. At each step we select the most promising solutions based on their performance $\mathcal{P}$ and we further develop them by using \texttt{GenerateCandidates}. In this way, at each step, the algorithm evaluates solutions of the same length. The number of solutions to be expanded at each step is the beam width $w$, and the algorithm terminates once the maximum allowed length is reached. In our experiment $w = \vert \mathcal{T} \vert$.

For Tabu search (TS) \cite{glover1989tabu, glover1990tabu}, we define the fitness function $\mathcal{F}_t$ for a candidate curriculum to be equal to $\mathcal{P}$. We start the search by randomly selecting a curriculum in $\mathcal{C}_{\leq L}$. At each iteration we perform local changes to the current best solution to generate the relative neighborhood. We first generate a list of curricula $R$ composed by all the curricula obtained by adding or removing a task from/to the last task in the current best curriculum. Then we generate all the curricula resultant of any pairwise swap of any two tasks of any curriculum in $R$. The size of our tabu list is $T$, and, when full, we empty it following a FIFO strategy. If during the search all the curricula in the neighborhood are in the tabu list the new current best solution is randomly selected. The algorithm terminates after a fixed number of iterations. In our experiments $T = 30$.

In Genetic Algorithm (GA) \cite{goldberg1989genetic}, we set the initial population as $U$ randomly sampled curricula from $\mathcal{C}_{\leq L}$ and, similarly to tabu search, we set the fitness function $\mathcal{F}_g$ for a candidate curriculum to be equal to $\mathcal{P}$. At each iteration of the genetic algorithm we select two parents from the current population $N_g$ with a roulette wheel selection. Given a candidate curriculum $c_i$, its probability of being selected as a parent is $p_i = \mathcal{F}_g(c_i)/\sum_{c \in N_g} \mathcal{F}_g(c)$.
Given two parent curricula we generate a new population of $U$ candidate curricula by applying a standard single point cross over at randomized lengths along each parent gene (sequence of tasks). Each cross over step produces two children curricula, and the process is repeated until $U$ children curricula are created.
We also included a form of elitism in order to improve the performance of the algorithm by adding the parents to the population they generated. Genetic algorithms also include the definition of a mutation operator. In our implementation this acts on each candidate curriculum in the newly generated population with probability $p_m$. The mutation can be of two equally probable types: task-wise mutation, which given a candidate curriculum of length $l$ changes each of its intermediate tasks with probability equal to $1/l$; length-wise mutation, where equal probability is given to either dropping or adding a new task at a randomly selected position of a candidate curriculum. In case of candidate curricula composed by one task only, the dropping option for the length-wise mutation does not apply. The algorithm terminates after a fixed number of iterations. In our experiments $U = 50$ and $p_m = 0.5$.

Ant Colony Optimization (ACO) \cite{dorigo1991ant} is a CO metaheuristic which consists in deploying multiple agents (ants), in the given search space for depositing artificial pheromone along the most successful trails in that space, thus guiding the search towards more and more successful solutions. Each agent starts from an empty sequence of tasks. At each step an agent moves towards the goal by adding a new  task to the current candidate curriculum $c$ which represents the trail walked by the ant. Given a task $m_i$ its probability of being selected is $P(m_i) = [(\tau_{m_i}+K)^\alpha + I_{m_i}^\beta] / [\sum_{E}{[(\tau_{m_j}+K)^\alpha + I_{m_j}^\beta}]]$ with $E = \{ m_j \in \mathcal{T} \vert m_j \notin c \}$.The visibility $I_{m_i}$ is calculated as the performance improvement obtained by adding task $m_i$ to the current candidate curriculum when positive, and zero otherwise. Parameters $\alpha$ and $\beta$ control the influence of the pheromone versus the improvement while $K$ is a threshold to control from what pheromone value the search starts to take it into account. Once the maximum curriculum length is reached, artificial pheromone is deposited on all curricula explored in this way, from the first to the best along the last trail, concluding in this way an iteration. The pheromone evaporation rate is specified with the parameter $\rho$ while the maximum level of pheromone to be accumulated over a candidate solution is set to $f_{max}$.  The algorithm terminates after a fixed number of iterations. In our experiments $\alpha = 1, \beta = 1.2, K = 5, f_{max} = 50, \rho = 0.2$ and the number of ants is $20$. The parameters of all algorithms have been fine-tuned manually across all experiments.

\section{Experimental Evaluation}

We performed a thorough experimental evaluation on two domains, with two sets of experiments each. Our goal is twofold. First, we intend to show that curriculum learning is indeed a valid method to improve regret, jumpstart, and max-return over learning from scratch, which makes it applicable to the new scenarios we consider. Second, we evaluate the algorithms introduced above, to establish which one is the most appropriate for curriculum learning. 

The two domains,
BlockDude and Gridworld, were implemented within the software library Burlap~\footnote{http://burlap.cs.brown.edu}. 
We used the implementation of Sarsa($\lambda$) available in Burlap with $\lambda = 0.9$, learning rate $\alpha = 0.1$ and discount factor $\gamma = 0.999$. We represented the action-value function through the Burlap Tile Coding function approximator, initializing the Q-values to $q(s,a) = 0$ for all state-action pairs. The agents explore with an $\epsilon$-greedy policy, with $\epsilon=0.1$ and decreasing linearly from $0.1$ to $0$ in the last $25\%$ of the episodes. 

\subsection{GridWorld}

GridWorld is an implementation of an episodic grid-world domain used in the evaluation of existing curriculum learning methods~\cite{leonetti17,daSilva2018}. Each cell can be free, or occupied by a  \emph{fire}, \emph{pit}, or \emph{treasure}. The agent can move in the four cardinal directions, and the actions are deterministic. The reward is $-2500$ for entering a pit, $-500$ for entering a fire, $-250$ for entering the cell next to a fire, and $200$ for entering a cell with the treasure. The reward is $-1$ in all other cases. The episodes terminate under one of these three
conditions: the agent falls into a pit, reaches the treasure, or executes a maximum number of actions (50). The variables fed to tile coding are the distance from the treasure (which is global and fulfills the Markov property), and distance from any pit or fire within a radius of $2$ cells from the agent (which are local variables, and allow the agent to learn how to deal with these objects when they are close, and transfer this knowledge).

\subsection{BlockDude}

BlockDude is another domain available in Burlap, which has also been used for curriculum learning~\cite{leonetti17}. It is a puzzle game where the agent has to stack boxes in order to climb over walls and reach the exit. The available actions are moving left, right, up, pick up a box and put down a box. The agent receives a reward of $-1$ for each action taken. The variables used as input to tile coding are distance
from the exit, distance from each box, distance from each edge of the map, direction of the agent (binary) and whether or not it is holding a box (also binary).

\begin{figure}[htbp]
	\centering
	\includegraphics[width=\linewidth]{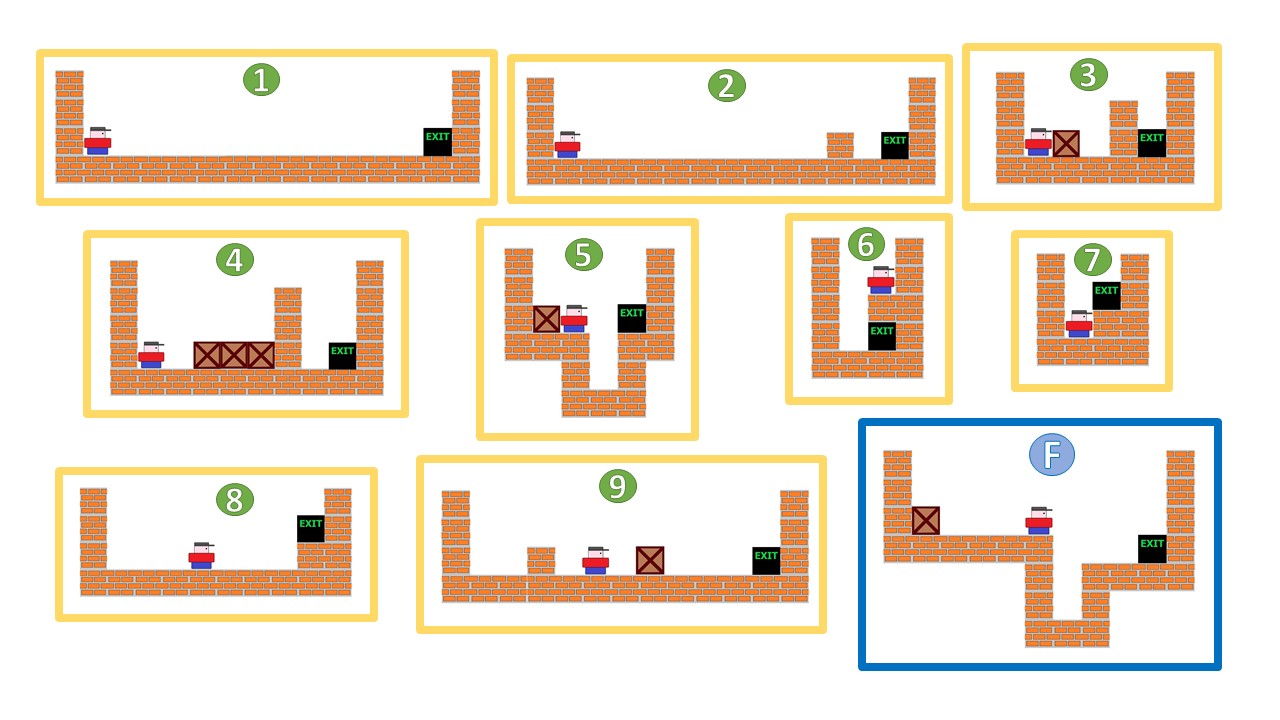}
	\caption{Intermediate (in yellow) and final task (in blue) of the
		second experiment in the BlockDude domain. The global optimum
		is the curriculum 7-1-5.} \label{Fig:Domains}
\end{figure}

\subsection{Transfer Learning}

We used an egocentric representation (using distances with respect to the agent) and local variables to favor transfer, as described separately for the two domains. We also normalized the variables in $[0,1]$, so that the input is invariant to the scale of the domain. Furthermore, we used a particular value-function transfer~\cite{tl4rl} inspired by Concurrent Layered Learning \cite{whiteson2003concurrent}. In Concurrent Layered Learning, an agent learns a complex behavior by incrementally learning sub-behaviors (layers). The more complex behaviors (higher layers) directly depend on the easier ones (lower layers), and during training all the layers are updated simultaneously.

This concept was implemented by carrying over the features of the source task into the target task, along with its parameters. Let $V_i$ be the set of variables defined for the source task, and $V_j$ the variables defined for the target task, in a source-target pair along the curriculum. Let $q_i(s,a) = \bm{\theta}^T_i \bm{\phi}_i(s,a)$ be the value function of the source task. The value function for the target task, $q_j$, is defined as:
\begin{equation*}
q_j(s,a) \coloneqq \begin{cases}
\bm{\theta}^T_i \bm{\phi}_i(s,a) + \bm{\theta}^T_j \bm{\phi}_j(s,a), & \text{if $V_i \subseteq V_j$}.\\
\bm{\theta}^T_j \bm{\phi}_j(s,a), & \text{otherwise}.
\end{cases}
\end{equation*}
where $\bm{\phi}_j$ is the feature vector of the target task. Therefore, if the variables of the target are compatible with the variables of the source, so that the features of the source are defined in the target, the features and their parameters are carried over. The new parameters, introduced in the target, are set to $0$ so that the imported features initially dominate the behavior. In our experiments we do not remove features, and the number of features and parameters grows with every transfer. However, it is possible to perform feature selection, and remove the features that do not affect the value function significantly.

\subsection{Results}

We chose two relatively small domains so that we could perform a thorough evaluation, by computing and analyzing all curricula within the given maximum length.

In our experiments, we perform an evaluation phase each $K=10$ episodes in order to estimate the quality of the learned policy. As the environments are deterministic, we can perform each evaluation step $Q=1$ times. Each curriculum has been executed $10$ times, and its value for each metrics estimated as the average over those trials.

We ran two sets of experiments per domain, one in which the number of tasks is high and the maximum length is low, and one in which, on the contrary, the number of tasks is low, but the maximum length is high. For GridWorld, the first set of experiments has parameters $n \coloneqq \lvert \mathcal{T} \rvert =12$ and $L=4$, while the second $n=7$, and $L=7$. For BlockDude, the first set of experiments has parameters $n=18$ and $L=3$, while the second $n=9$ and $L=5$. These parameters were chosen so that the total number of curricula does not exceed $20000$. For both domains, the intermediate tasks have been generated manually, by varying the size of the environment, adding and removing elements (pits and fires in GridWorld, and columns and movable blocks in BlockDude). Figure \ref{Fig:Domains} shows all the intermediate tasks, and the final task (marked with F) for one of the two BlockDude experiments. All experiments have a different final task, and set of intermediate tasks.
All tasks are run for a number of episodes that ensures that the agent has converged to the optimal policy, and were determined at the time of task generation.

\begin{table*}[htbp]
	\centering
	\resizebox{\linewidth}{!}{
	\midsepremove
		\begin{tabular}{c|c|c|c|c|c|c|c|c|c|c|c|c|c|c}
			\toprule
			{} &  \multicolumn{7}{c|}{\cellcolor{Red}n = 12; L = 4; tot = 13345}& \multicolumn{7}{c}{\cellcolor{Yellow}n = 7; L = 7; tot = 13700}\\
			\midrule
			GW   & $C_0$ & Greedy   & GA & Tabu  & ACO & Beam   & Opt & $C_0$ & Greedy   & GA & Tabu  & ACO & Beam   & Opt\\
			\bottomrule
			\multirow{3}{*}{REG} 
			&-&24&378.58&\cellcolor{Green}\textbf{364.54}&378&373&-&-&16&155.84&\cellcolor{Green}\textbf{165.4}&168&155&-\\
			&-0.36&-0.26&-0.23&\cellcolor{Green}\textbf{-0.21}&-0.24&-0.22&-0.19&-0.49&0.43&-0.41&\cellcolor{Green}\textbf{-0.38}&-0.3887&-0.39&-0.28\\
			&-&-&[-0.24:-0.23]&\cellcolor{Green}\textbf{[-0.21:-0.21]}&[-0.24:-0.24]&-&-&-&-&[-0.42:-0.39]&\cellcolor{Green}\textbf{[-0.38:-0.37]}&[-0.40:-0.38]&-&-\\

			\bottomrule
			
			\multirow{3}{*}{JS} 
			&-&34&378.7&380.94&378&\cellcolor{Green}\textbf{373}&-&-&16&156.22&169.9&\cellcolor{Green}\textbf{154}&155&-\\
			&-2283.8&-860.03&-1223.24&-842.68&-779.82&\cellcolor{Green}\textbf{-738.29}&-601.74&-2624&-827.68&-896.17&-855.99&\cellcolor{Green}\textbf{-725.65}&-773.71&-360.9\\
			&-&-&[-1309.87:-1136.61]&[-900.18:-785.17]&[-795.44:-764.21]&\cellcolor{Green}-&-&-&-&[-966.24:-826.11]&[-920.08:-791.97]&\cellcolor{Green}\textbf{[-759.65:-691.65]}&-&-\\
			
			\bottomrule
			\multirow{3}{*}{TTT} 
			&-&24&380.3&381.86&39&\cellcolor{Green}\textbf{373}&-&-&12&156.52&166.3&168&\cellcolor{Green}\textbf{155}&-\\
			&-1351.6&-315.6&-545.73&-315.6&-466.12&\cellcolor{Green}\textbf{-315}.6&-315.6&-2788.1&-1643.7&-2300.41&-19880.42&-1716.13&\cellcolor{Green}\textbf{-1643.87}&-1643.87\\
			&-&-&[-388.93:-702.53]&[-315.6:-315.6]&[-345.80:-586.44]&\cellcolor{Green}-&-&-&-&[-1967.20:-2633.63]&[-17369.97:-22390.87]&[-1680.85:-1751.42]&\cellcolor{Green}-&-\\
			
			\bottomrule
			
			\multirow{3}{*}{MR} 
			&-&24&377.16&463.28&378&\cellcolor{Green}\textbf{373}&-&-&19&155.98&\cellcolor{Green}\textbf{165.92}&168&155&-\\
			&-50&-26.6&-15.5&-5.716&-5.33&\cellcolor{Green}\textbf{-3}&19.9&-50&21.1&6.96&\cellcolor{Green}\textbf{26.78}&18.36&21.1&68.5\\
			&-&-&[-18.55:-12.45]&[-9.87:1.56]&[-6.90:-3.76]&\cellcolor{Green}-&-&-&-&[0.95:12.98]&\cellcolor{Green}\textbf{[22.84:30.71]}&[14.52:22.21]&-&-\\
			\bottomrule
		\end{tabular}
	\midsepdefault
	}
	\caption{ 
		Results of all the algorithms on the GridWorld domain. Green cells enhance the result of the best algorithm for a specific performance and experiment.}
	\label{Table:GridWorld}
\end{table*}
\begin{table*}[htbp]
	\centering
	\resizebox{\linewidth}{!}{
		\midsepremove
		\begin{tabular}{c|c|c|c|c|c|c|c|c|c|c|c|c|c|c}
			\toprule
			{} &  \multicolumn{7}{c|}{\cellcolor{Red}n = 9; L = 5; tot = 18730} & \multicolumn{7}{c}{\cellcolor{Yellow}n = 18; L = 3; tot = 5221}\\
			\midrule
			BD   & $C_0$ & Greedy   & GA & Tabu  & ACO & Beam   & Opt & $C_0$ & Greedy   & GA & Tabu  & ACO & Beam   & Opt\\
			\bottomrule
			\multirow{3}{*}{REG}
			&-&25&266.78&286.02&245&\cellcolor{Green}\textbf{244}&-&-&52&622.66&750.5&663&\cellcolor{Green}\textbf{613}\\
			&-0.81&-0.45&-0.45&-0.45&-0.46&\cellcolor{Green}\textbf{-0.40}&-0.35&-0.80&-0.52&-0.50&-0.45&-0.53&\cellcolor{Green}\textbf{-0.43}&-0.42\\
			&-&-&[-0.47:-0.43]&[-0.47:-0.43]&[-0.49:-0.43]&\cellcolor{Green}-&-&-&-&[-0.52:-0.49]&[-0.45:-0.44]&[-0.54:-0.51]&\cellcolor{Green}-&-\\
			
			\bottomrule
			
			\multirow{3}{*}{JS} 
			&-&25&267.44&315.34&245&\cellcolor{Green}\textbf{244}&-&-&52&623.66&\cellcolor{Green}\textbf{695.16}&663&613\\
			&-49.9&-49.04&-49.08&-48.17&-48.42&\cellcolor{Green}\textbf{-47.32}&-47.77&-50&-49.4&-49.66&\cellcolor{Green}\textbf{-49.36}&-49.76&-49.38&-49\\
			&-&&[-49.18:-48.98]&[-48.58:-47.77]&[-48.75:-48.09]&\cellcolor{Green}-&-&-&-&[-49.70:-49.62]&\cellcolor{Green}\textbf{[-49.41: -49.32]}&[-49.79:-49.73]&-&-\\

			\bottomrule
			\multirow{3}{*}{TTT} 
			&-&18&267.92&262.86&245&\cellcolor{Green}\textbf{244}&-&-&36&625.28&665.2&663&\cellcolor{Green}\textbf{613}\\
			&-2981.7&-2749.3&-2879.87&-3086.52&-2793.00&\cellcolor{Green}-\textbf{2749.3}&-2749.3&-4986&-3828&-3868.10&-3828&-3834.80&\cellcolor{Green}-\textbf{-3828}&-3828\\
			&-&&[-2825.41:-2934.32]&[-2998.12:-3174.92]&[-2768.23:-2817.78]&\cellcolor{Green}-&-&-&-&[-3852.95:-3883.25]&[-3828:-3828]&[-3831.12:-3838.47]&\cellcolor{Green}-&-\\
			
			\bottomrule
			
			\multirow{3}{*}{MR} 
			&-&31&266.66&333.76&245&\cellcolor{Green}\textbf{244}&-&-&67&621.34&752.1&663&\cellcolor{Green}\textbf{613}&-\\
			&-10&-10&-10&-10&-10&\cellcolor{Green}\textbf{-10}&-10&-14&-14&-14&-14&-14&\cellcolor{Green}\textbf{-14}&-14\\
			&-&-&[-10:-10]&[-10:-10]&[-10:-10]&\cellcolor{Green}-&-&-&-&[-14:-14]&[-14:-14]&[-14:-14]&\cellcolor{Green}-&-\\

			\bottomrule
		\end{tabular}
	\midsepdefault
	}
	\caption{ 
		Results of all the algorithms on the BlockDude domain. Green cells enhance the result of the best algorithm for a specific performance and experiment.}
	\label{Table:BlockDude}
\end{table*}

In Table \ref{Table:GridWorld} and \ref{Table:BlockDude} we show the results of the experiments on GridWorld and BlockDude respectively. The header row shows the number of candidate tasks, maximum length, and the total number of curricula of each experiment. Each cell contains the average number of curricula evaluated, the value of the objective function, and the confidence interval of that value, for the corresponding metric. We also included the optimal curriculum (last column of each experiment) and learning with no curriculum (denoted as $C_0$). The regret is normalized in $[-1,0]$ with $0$ being the value of the policy achieving no regret. Since we are maximizing, for all metrics a larger value is always better. Our implementation of Tabu Search, Genetic Algorithm and Ant Colony are stochastic and non-terminating, so that in the limit they always find the optimal solution. For a fair comparison, and for the cases of practical interest, we interrupted them at the first iteration (for instance, generation in GA) in which they evaluated a number of curricula equal to or greater than Beam Search at its optimal value, since Beam Search is deterministic.

From tables \ref{Table:GridWorld} and \ref{Table:BlockDude} it results clear how Curriculum Learning can be employed for optimizing different performance metrics for learning the final task. Moreover curricula that scored well for one metric, often showed poor performance in the other metrics. An important consequence of this is the need of powerful optimization algorithms able to sensibly choose which source tasks should be sequenced to become the curriculum optimizing a specific performance over the final task. From our study, Beam Search outperforms all the metaheuristics in almost all the experiments, and when it does not, the solution is often close to the best one. Tabu Search, notably the other trajectory-based algorithm, is better in some cases. However, since Beam Search is deterministic, it may be preferable as there is no variation between runs. It is also important to notice that the performance of Beam Search degrades faster as the maximum length increases, rather than as the number of candidate tasks increases. 
Lastly, time-to-threshold, the objective used the most in the literature, also appears to be the easiest to optimize. On the other hand, regret seems to be the most difficult metric. 

BlockDude appears to be too simple for max-return to be a viable objective, since all curricula, including no curriculum, eventually learn the optimal policy. On the contrary, in GridWorld, different curricula result in the discovery of different policies, with a significant improvement over learning from scratch.


\section{Conclusion}
We introduced a framework for task sequencing in curriculum learning, proposing two novel scenarios which led to the adoption of three metrics in addition to time-to-threshold, 
the one mostly used in the literature. With the exception of regret minimization, which has been used to explore efficiently in single tasks, the other metrics are directly borrowed from general transfer learning metrics. Since transfer learning is a fundamental component of curriculum learning, this is a reasonable first step. However, it is possible that metrics specific to curriculum learning will be designed in future work.

We adapted four metaheuristc algorithms to the problem of task sequencing, and evaluated them on four experiments. Trajectory-based methods outperformed population based method, evaluating a small fraction of the total number of curricula. We demonstrated that curriculum learning can be used not only to learn faster, but also explore more efficiently (by maximizing jumpstart and regret) and learn better policies (by maximizing max-return) than learning from scratch. While metaheuristc algorithms are fairly general, we expect that even better results can be obtained by developing specific heuristic algorithms tailored for a particular objective function.


\bibliographystyle{IEEEtran}
\bibliography{bibfile}

\end{document}